\documentclass[conference,a4paper]{IEEEtran}
\IEEEoverridecommandlockouts

\usepackage[hidelinks]{hyperref}
\usepackage[cmex10]{amsmath}
\usepackage{amssymb,amsfonts}
\usepackage{multirow}
\usepackage[ruled,vlined]{algorithm2e}
\usepackage{graphicx}
\graphicspath{{Figures/PDF/}{Figures/PNG/}}
\usepackage{subcaption}  
\usepackage{float}
\usepackage{booktabs}
\usepackage{siunitx}
\usepackage[numbers,compress]{natbib}
\usepackage{texnames}
\usepackage{bm,bbm}
\usepackage{orcidlink}

\begin{document}

\title{\uppercase{3D Semantic Segmentation for Post-Disaster Assessment}
}

\author{	\IEEEauthorblockN{Nhut Le}
	\IEEEauthorblockA{\textit{Lehigh University}\\
		Bethlehem, PA 18015, USA\\
		nhl224@lehigh.edu}
	\and
	\IEEEauthorblockN{Maryam Rahnemoonfar}
	\IEEEauthorblockA{\textit{Lehigh University}\\
		Bethlehem, PA 18015, USA\\
		maryam@lehigh.edu}
}

\maketitle
\begin{abstract}
The increasing frequency of natural disasters poses severe threats to human lives and leads to substantial economic losses. While 3D semantic segmentation is crucial for post-disaster assessment, existing deep learning models lack datasets specifically designed for post-disaster environments. To address this gap, we constructed a specialized 3D dataset using unmanned aerial vehicles (UAVs)-captured aerial footage of Hurricane Ian (2022) over affected areas, employing Structure-from-Motion (SfM) and Multi-View Stereo (MVS) techniques to reconstruct 3D point clouds. We evaluated the state-of-the-art (SOTA) 3D semantic segmentation models, Fast Point Transformer (FPT), Point Transformer v3 (PTv3), and OA-CNNs on this dataset, exposing significant limitations in existing methods for disaster-stricken regions. These findings underscore the urgent need for advancements in 3D segmentation techniques and the development of specialized 3D benchmark datasets to improve post-disaster scene understanding and response.

\end{abstract}


\section{Introduction}
\begin{figure*}[!t]
\centering
\begin{subfigure}{0.45\linewidth}
\includegraphics[width=.9\linewidth]{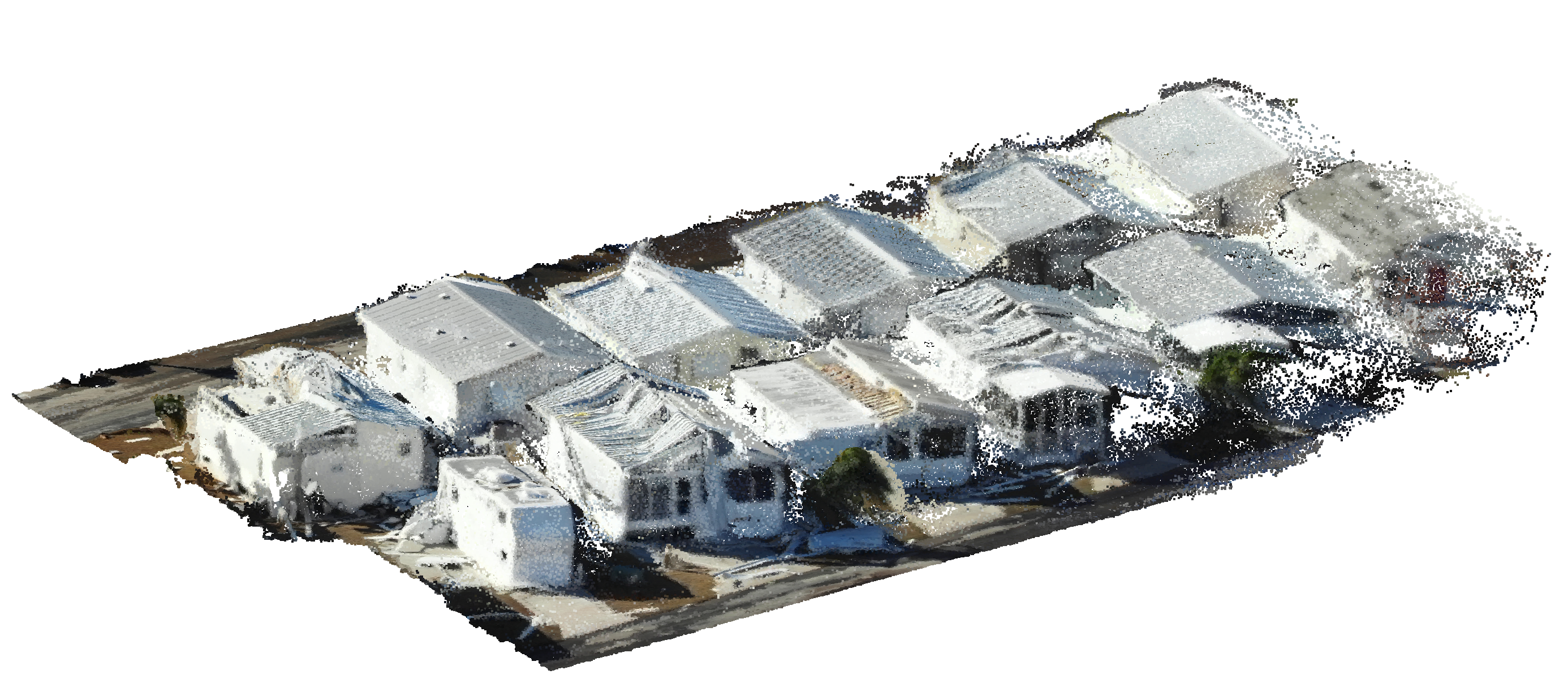}
\end{subfigure}
\begin{subfigure}{0.45\linewidth}
\includegraphics[width=.9\linewidth]{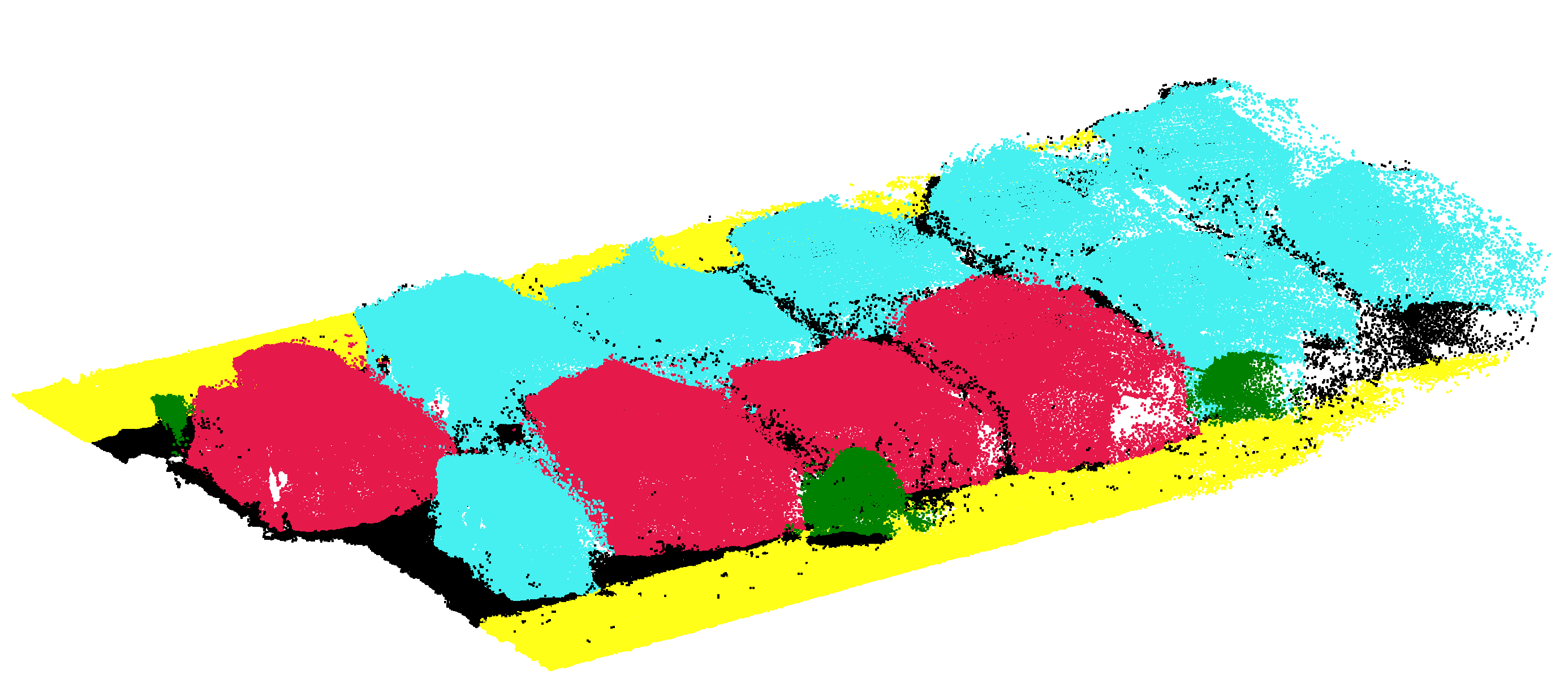}
\end{subfigure}
\begin{subfigure}{0.45\linewidth}
\includegraphics[width=.9\linewidth]{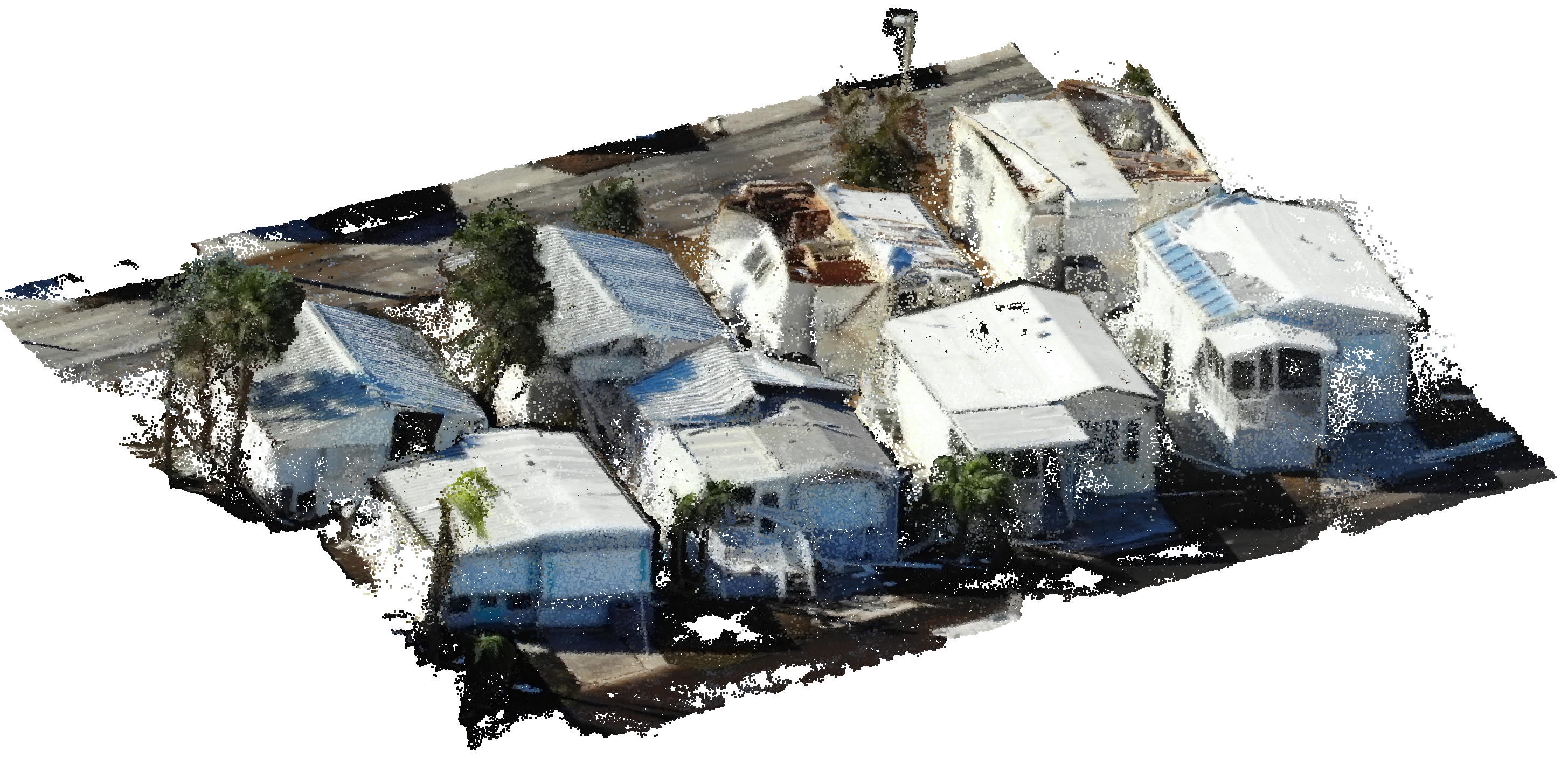}
\end{subfigure}
\begin{subfigure}{0.45\linewidth}
\includegraphics[width=.9\linewidth]{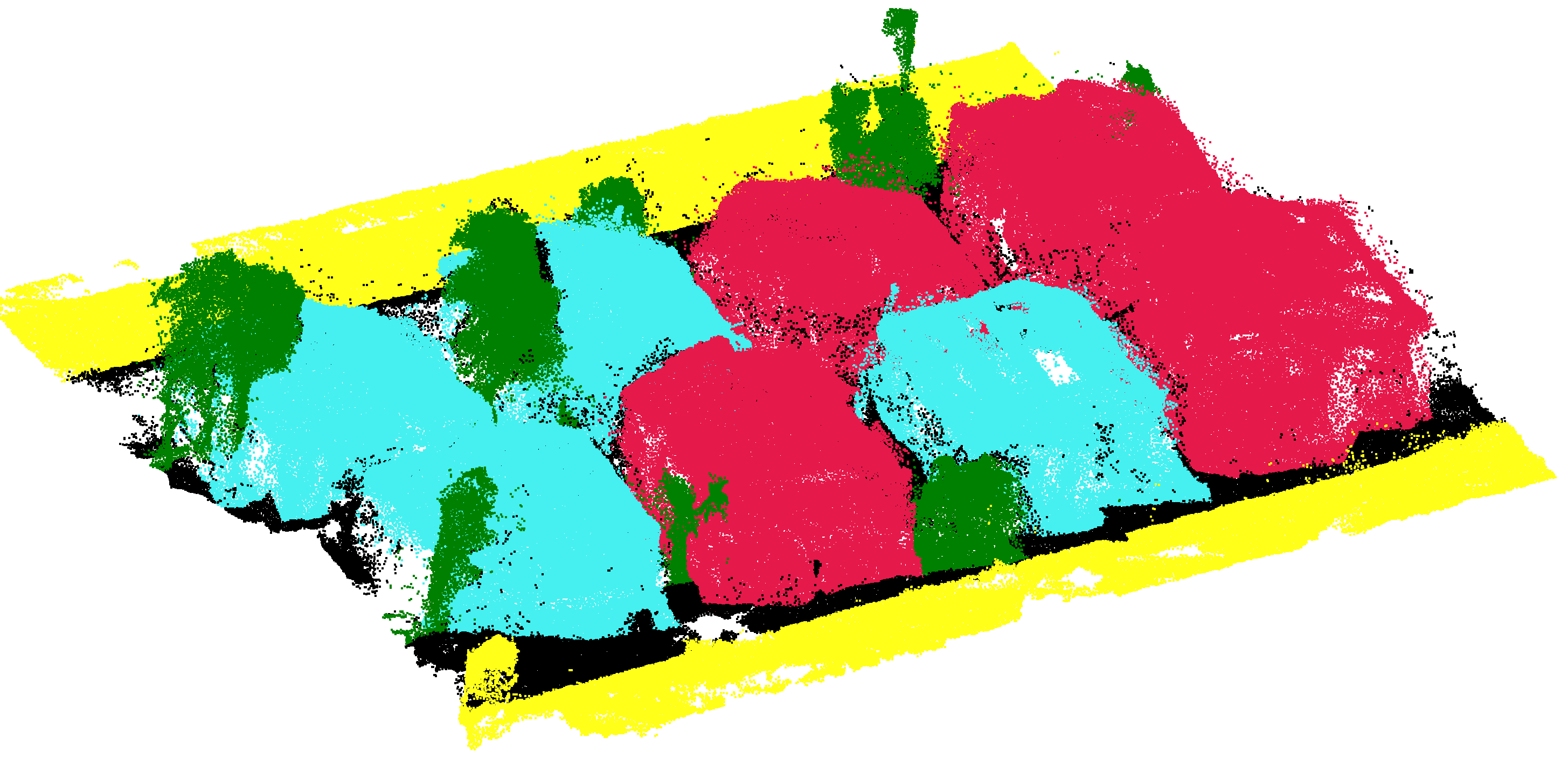}
\end{subfigure}
\caption{\textbf{Point clouds} in our dataset with semantic labels: \textbf{Cyan} – Building (no damage), \textbf{Red} – Building (damaged), \textbf{Yellow} – Road, \textbf{Green} – Tree, \textbf{Black} – Background}
\label{fig:train}
\end{figure*}
In recent years, the frequency and severity of natural disasters have escalated, posing significant risks to human lives and causing substantial economic losses \cite{undrr2020human}. To mitigate these challenges, emerging technologies such as autonomous vehicles, robotics, and virtual/augmented reality offer promising solutions for enhancing rescue operations, reducing operational hazards, and improving post-disaster assessment. A key component in these advancements is the accurate interpretation of 3D point clouds, where 3D semantic segmentation plays a crucial role.
\\
Recent deep learning-based approaches \cite{qi2017pointnet, qi2017pointnetplusplus, NEURIPS2018_f5f8590c, Thomas_2019_ICCV, hu2019randla, zhao2021point, wu2022point, Peng_2024_CVPR, wu2024ptv3} have demonstrated state-of-the-art (SOTA) performance on high-quality indoor \cite{armeni2016s3dis, dai2017scannet, rozenberszki2022language, yeshwanthliu2023scannetpp} and outdoor \cite{behley2019iccv, 9150622, Sun_2020_CVPR, Caesar_2020_CVPR, Liao2022PAMI, gaydon2024fractalultralargescaleaeriallidar} 3D benchmark datasets. However, to the best of our knowledge, no existing 3D benchmark datasets or deep learning-based 3D semantic segmentation methods have been specifically developed for post-disaster scenarios.
\\
To address this gap, we reconstructed a specialized 3D dataset tailored for disaster-stricken environments and evaluated the performance of the SOTA 3D deep learning-based semantic segmentation methods, Fast Point Transformer (FPT) \cite{park2022fast}, Point Transformer v3 (PTv3)\cite{wu2024ptv3}, and OA-CNNs \cite{Peng_2024_CVPR}, on this dataset. Inspired by RescueNet \citep{rahnemoonfar2023rescuenet}, we collected aerial footage of areas affected by Hurricane Ian (2022) in Florida using unmanned aerial vehicles (UAVs). Using Structure-from-Motion (SfM) \cite{schoenberger2016sfm} and Multi-View Stereo (MVS) \cite{schoenberger2016mvs} techniques, we reconstructed 3D point clouds from the captured footage. Ground-truth 3D semantic segmentation labels were generated through manual annotation of 2D images.
\\
To assess the capabilities of FPT \cite{park2022fast}, PTv3 \cite{wu2024ptv3}, and OA-CNNs \cite{Peng_2024_CVPR} in post-disaster scenarios, we conducted experiments on our reconstructed 3D dataset. The evaluation revealed significant limitations in existing methods when applied to post-disaster environments, highlighting the urgent need for advancements in 3D semantic segmentation techniques and the creation of specialized benchmark datasets to better address the unique complexities of post-disaster scenarios.

\section{Related Work}
\subsection{Datasets and Benchmarks for Damage Assessment}
The xBD dataset \cite{gupta2019xbddatasetassessingbuilding} is a large-scale collection of pre- and post-disaster satellite imagery designed for building damage assessment. It categorizes buildings based on damage levels, such as no damage, minor, major, and destroyed, while also tagging environmental factors like floodwater and fire in specific scenes. Similarly, SpaceNet \cite{9857340} serves as a satellite benchmark focused on flood scenarios, combining tasks such as building footprint detection, road network extraction, and flood segmentation.
In addition to satellite imagery, aerial and UAV (drone) imagery datasets have been developed for higher-resolution analysis. One such dataset, FloodNet \cite{rahnemoonfar2020floodnethighresolutionaerial}, contains UAV images from Hurricane Harvey, with pixel-level annotations for nine classes, including flooded vs. non-flooded roads, buildings, water, and debris. FloodNet was among the first to offer multi-class semantic annotations in a post-hurricane context, extending beyond just buildings. More recently, RescueNet \cite{rahnemoonfar2023rescuenet} gathered 4,494 UAV images after Hurricane Michael, providing comprehensive annotations across 10 classes. RescueNet includes semantic segmentation masks for buildings, roads (clear or blocked), trees, vehicles, pools, and water, along with a four-tier damage level scale for each building (superficial, minor, major, total destruction).
\subsection{3D Benchmark Datasets}
3D semantic segmentation datasets are created using RGB-D cameras, LiDAR scanners, or synthetic 3D data from rendering engines like Unreal Engine. They are categorized into indoor and outdoor datasets.
\subsubsection{3D Indoor Benchmark Datasets}
S3DIS \citep{armeni2016s3dis}, built with an RGB-D scanner, captures point clouds of various indoor spaces. The ScanNet family (ScanNet \citep{dai2017scannet}, ScanNet200 \citep{rozenberszki2022language}, ScanNet++ \citep{Yeshwanth_2023_ICCV}) reconstructs 3D point clouds from RGB-D video sequences using fusion algorithms, with semantic annotations via Amazon Mechanical Turk.

\subsubsection{3D Outdoor Benchmark Datasets}
SemanticKITTI \citep{behley2019iccv}, from the KITTI Vision Odometry Benchmark, features 22 urban driving sequences collected with a Velodyne HDL-64E LiDAR. Waymo Open Dataset \citep{Sun_2020_CVPR} covers multiple cities using autonomous vehicles (AVs) equipped with LiDAR and high-resolution cameras. nuScenes \citep{Caesar_2020_CVPR} provides 1,000 urban scenes from Boston and Singapore with multi-sensor AV data, including 3D bounding boxes for 23 object classes and attributes like moving and parked states.

\section{Methodology}
\subsection{3D Semantic Segmentation Methods}
\begin{figure*}[t!]
\centering
\includegraphics[width=\linewidth]{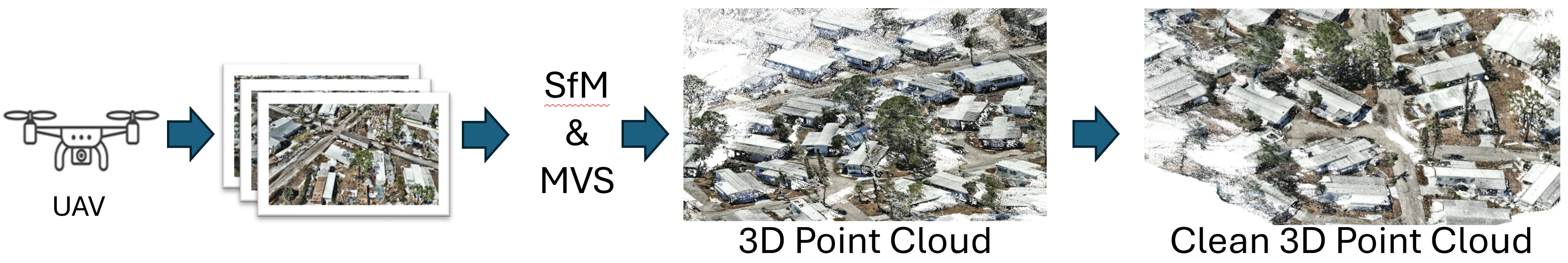}
\caption{\textbf{The reconstruction pipeline.} Frames extracted from UAV aerial footage are processed using Structure-from-Motion (SfM) and Multi-View Stereo (MVS) to generate 3D point clouds. Outliers are manually removed to obtain clean and accurate reconstructions.}
\label{fig:recon}
\end{figure*}



We evaluate the performance of three state-of-the-art (SOTA) models for 3D semantic segmentation: Fast Point Transformer (FPT) \citep{park2022fast}, Point Transformer V3 (PTv3) \citep{wu2024ptv3}, and OA-CNNs \citep{Peng_2024_CVPR}.
\\
Transformer-based architectures have demonstrated outstanding performance in natural language processing and 2D vision tasks such as image classification and object detection. However, their application to 3D point clouds is nontrivial due to the unordered and irregular structure of point cloud data, which contrasts with the grid-like structure expected by standard self-attention mechanisms.
\\
\textbf{FPT} \citep{park2022fast} addresses these computational challenges by leveraging voxelization to structure point clouds. It introduces centroid-aware voxelization and devoxelization techniques that preserve continuous coordinate embeddings, reduce quantization artifacts, and enhance coherence in dense predictions. These innovations significantly improve both the efficiency and accuracy of point cloud processing, especially for large-scale 3D scenes.
\\
\textbf{PTv3} \citep{wu2024ptv3} mitigates the limitations of conventional transformers in point cloud analysis by introducing a serialization strategy. It divides the 3D space into a uniform voxel grid and applies space-filling curves, such as Z-order, Hilbert, Trans Z-order, and Trans Hilbert curves, to convert unordered point sets into structured sequences while preserving spatial locality. This serialization enables attention mechanisms to operate effectively without relying on expensive neighbor search operations (e.g., k-nearest neighbors), as required in earlier models like PTv1 and PTv2 \citep{zhao2021point, wu2022point}. Moreover, PTv3 replaces complex relative positional encodings with a lightweight pre-positive sparse convolutional layer (xCPE), which captures spatial relationships efficiently with minimal computational cost.
\\
\textbf{OA-CNNs} \cite{Peng_2024_CVPR} tackle the adaptability limitations of traditional sparse convolutional networks, which often fail to capture varying local geometries in 3D scenes. To address this, OA-CNNs introduce two key modules: (1) \textbf{Adaptive Receptive Fields}, which dynamically adjust the receptive field size according to local geometric complexity, and (2) \textbf{Adaptive Relation Mapping}, which establishes dynamic interactions between features without the need for attention mechanisms. These modules enable OA-CNNs to capture both fine-grained geometric details and broad contextual patterns, achieving performance on par with or superior to transformer-based approaches such as PTv2 \citep{wu2022point}.

\subsection{Our 3D Dataset}

\subsubsection{3D Reconstruction}
To evaluate the performance of FPT \citep{park2022fast}, PTv3 \citep{wu2024ptv3}, and OA-CNNs \citep{Peng_2024_CVPR} in 3D semantic segmentation for the post-disaster scenario, we reconstructed a small 3D outdoor dataset. The reconstruction process consists of two main steps: data collection and 3D point cloud reconstruction from the collected data.

\begin{itemize} 
\item \textbf{Data Collection:} Our dataset was collected in the aftermath of Hurricane Ian, which impacted Florida in 2022. We obtained two extensive aerial footage sequences using UAVs flying over the affected areas. The aerial footage was then processed to extract individual frames for 3D reconstruction. To enhance the quality of the reconstruction, we applied preprocessing techniques, such as increasing contrast, to each frame to improve feature detection. 
   \item \textbf{3D Reconstruction:} We use Structure-from-Motion (SfM) \citep{schoenberger2016sfm} and Multi-View Stereo (MVS) \citep{schoenberger2016mvs} to reconstruct dense 3D point clouds from processed image frames. SfM generates a sparse 3D reconstruction and estimates camera poses through three steps: (1) detecting keypoints and extracting feature descriptors (e.g., SIFT \cite{sift}), (2) matching image pairs with RANSAC-based geometric verification, and (3) incrementally reconstructing the scene using triangulation and bundle adjustment. MVS builds on the sparse reconstruction to generate a dense point cloud by estimating depth maps from multiple overlapping views and refining them using photo-consistency and geometric constraints.
\end{itemize}
The dense point clouds were then manually cleaned to remove outliers, ensuring a more accurate dataset. As a result, we generated ten dense point clouds with an average number of points of 775000 across two affected areas.  Figure \ref{fig:recon} illustrates the pipeline used in reconstructing our 3D dataset.

\subsubsection{Ground Truth Generation}
\begin{figure*}[!t]
	\centering
	\includegraphics[width=\linewidth]{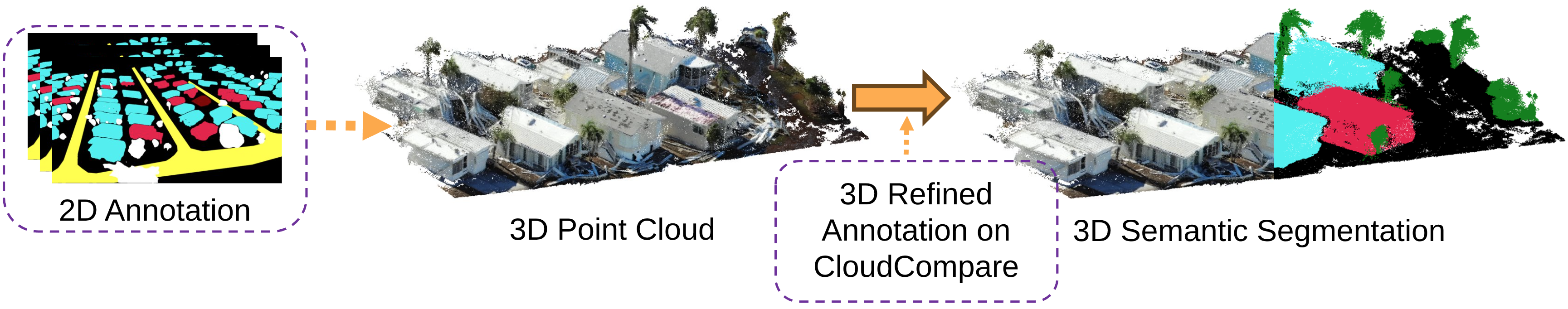}
	\caption{\textbf{3D Sem. Seg. Generation}. 2D annotations were projected into 3D space using majority voting across multiple frames, assigning each point the most frequent label from its 2D projections. Final refinements were made in CloudCompare \cite{cloudcompare} to ensure accurate point-level segmentation.}\label{fig:semantic}
\end{figure*}
To train deep learning-based methods effectively, generating accurate ground truth data is essential. In our 3D dataset, we employed a two-step process for ground truth generation: annotating 2D images and subsequently generating 3D semantic segmentation labels from these annotations.

\begin{itemize}
    \item \textbf{Annotation:} To reduce labor, we manually annotated every 10th frame in the collected data. Following the approach of RescueNet \cite{rahnemoonfar2023rescuenet}, we classified objects into distinct instance types and semantic classes, as outlined in Table \ref{tab:instance}.
    \begin{table*}[t!]
    \centering
    \caption{Instance types and classes in our 3D benchmark dataset} 
\label{tab:instance}
\begin{tabular}{|c|l|l|}
\hline
\textbf{Instance Type}    & \multicolumn{1}{c|}{\textbf{Classes}} & \multicolumn{1}{c|}{\textbf{Definition}}          \\ \hline
\multirow{2}{*}{Building} & Building-no-damage                    & Structures with little to no observable structural damage  \\ \cline{2-3} 
                          & Building-damage                 & Structures showing clear signs of structural damage   \\ \hline
                           Road                            & Road & Any type of road surface, regardless of condition 
                                  \\ \hline
Tree                      & Tree                                  & Isolated trees or tree clusters                          \\ \hline
Background                & Background                            & All other elements not categorized above, including terrain and miscellaneous objects                             \\ \hline
\end{tabular}
\end{table*}
    \item \textbf{3D Semantic Segmentation:} We generate 3D labels by aggregating the most frequent 2D labels for each point via 2D–3D correspondences using majority voting. This approach ensures consistent labeling across views. To correct potential noise or misclassifications in the ground truth, we manually refine the results using 3D editing software \cite{cloudcompare}.
    \end{itemize}
Figure \ref{fig:semantic} illustrates the pipeline used in generating our 3D ground truth.

\subsubsection{Dataset Splits} 

We divided the dataset into training, validation, and testing subsets. The training set comprises seven point clouds along with their associated 3D semantic segmentation labels. Both the validation and testing sets share three point clouds and their corresponding 3D semantic segmentations. Figure \ref{fig:train} presents two point clouds and their associated 3D semantic segmentations from the training set.

\section{Experiment and Result}

\subsection{Metric}
The performance of 3D semantic segmentation algorithms is commonly assessed using Overall Accuracy (OA), Mean Class Accuracy (mAcc), and Mean Intersection over Union (mIoU) on benchmark datasets. OA measures overall classification correctness, mAcc is the average per-class accuracy, and mIoU measures the overlap between predicted and ground truth regions.

\subsection{Evaluation our 3D dataset}
In our experiments, we utilized a computing environment equipped with an Intel Xeon Gold 6430 CPU and an NVIDIA RTX A5000 GPU with 24 GB GDDR6 memory. We trained FPT \citep{park2022fast}, PTv3 \citep{wu2024ptv3}, and OA-CNNs \citep{Peng_2024_CVPR} using the same configuration as for the S3DIS \citep{armeni2016s3dis} dataset, training from scratch.
\subsubsection{Quantitative Result}
As shown in Table~\ref{tab:res1}, all three methods yield low mIoU and mAcc scores. These results suggest that, despite their strong performance on existing 3D benchmark datasets, their effectiveness significantly declines when applied to our dataset, which features outdoor point clouds specific to post-disaster environments.
\begin{table}[ht]
\caption{\textbf{mIoU} and \textbf{mAcc} of of FPT \cite{park2022fast}, PTv3 \cite{wu2024ptv3}, and OA-CNNs \cite{Peng_2024_CVPR}on our dataset}
\centering
\begin{tabular}{|c|c|c|}
\hline
\textbf{Method}  & \textbf{mIoU}   & \textbf{mAcc}  \\ \hline
FPT   & 0.41         &       0.55       \\ \hline
PTv3      &      0.50       &   0.73         \\ \hline
OA-CNNs & 0.47 & 0.6 \\ \hline
\end{tabular}
\label{tab:res1}
\end{table}
\subsubsection{Qualitative Result}
As shown in Table~\ref{tab:res3}, there is a noticeable misclassification in the \textbf{Building-Damage} class, while the \textbf{Building-no-Damage} class is generally well labeled. This indicates that all three models struggle to distinguish between damaged and undamaged buildings, likely due to the subtle texture differences in post-disaster scenarios.
{
\setlength{\tabcolsep}{1pt}
\begin{table}[!t]
\caption{\textbf{Qualitative Results.} The outputs of FPT \cite{park2022fast}, PTv3 \cite{wu2024ptv3}, and OA-CNNs \cite{Peng_2024_CVPR} on the test set. Building-no-Damage (Cyan), Building-Damage (Red), Tree (Green), Road (Yellow), and Background (Black)}
\centering
\begin{tabular}{c|c}
\textbf{Input}  &  \includegraphics[width=0.7\linewidth]{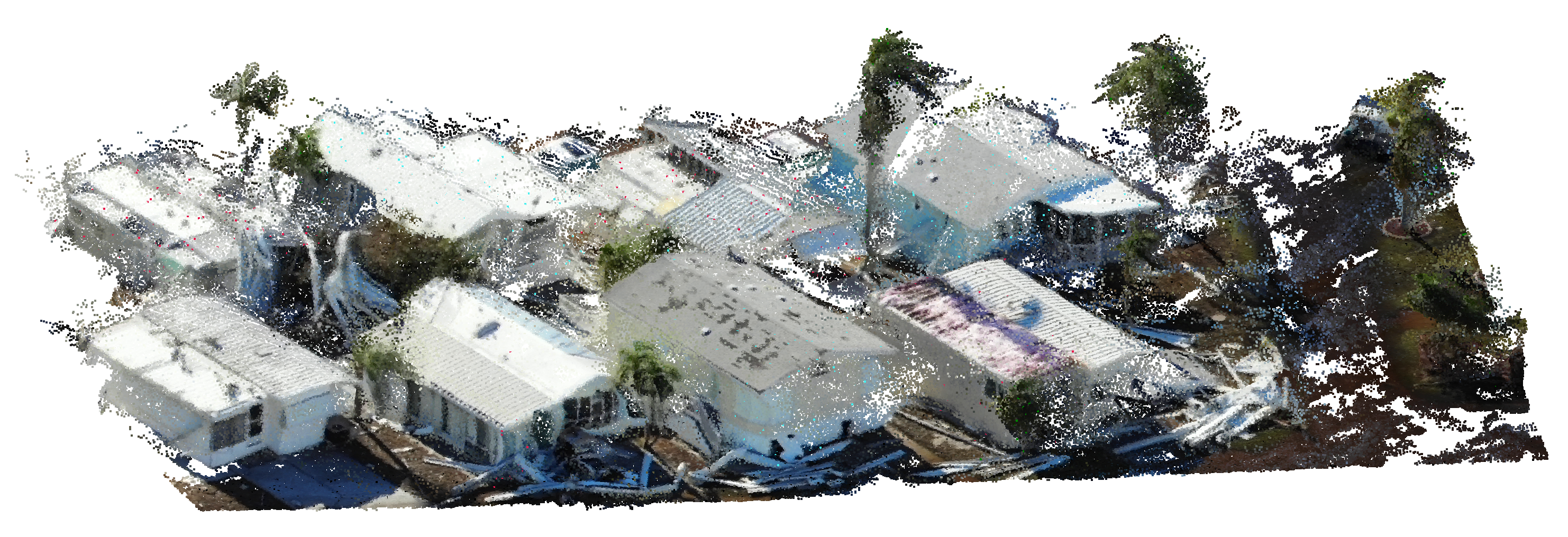} \\ \hline
\textbf{Ground Truth}
  & \includegraphics[width=0.7\linewidth]{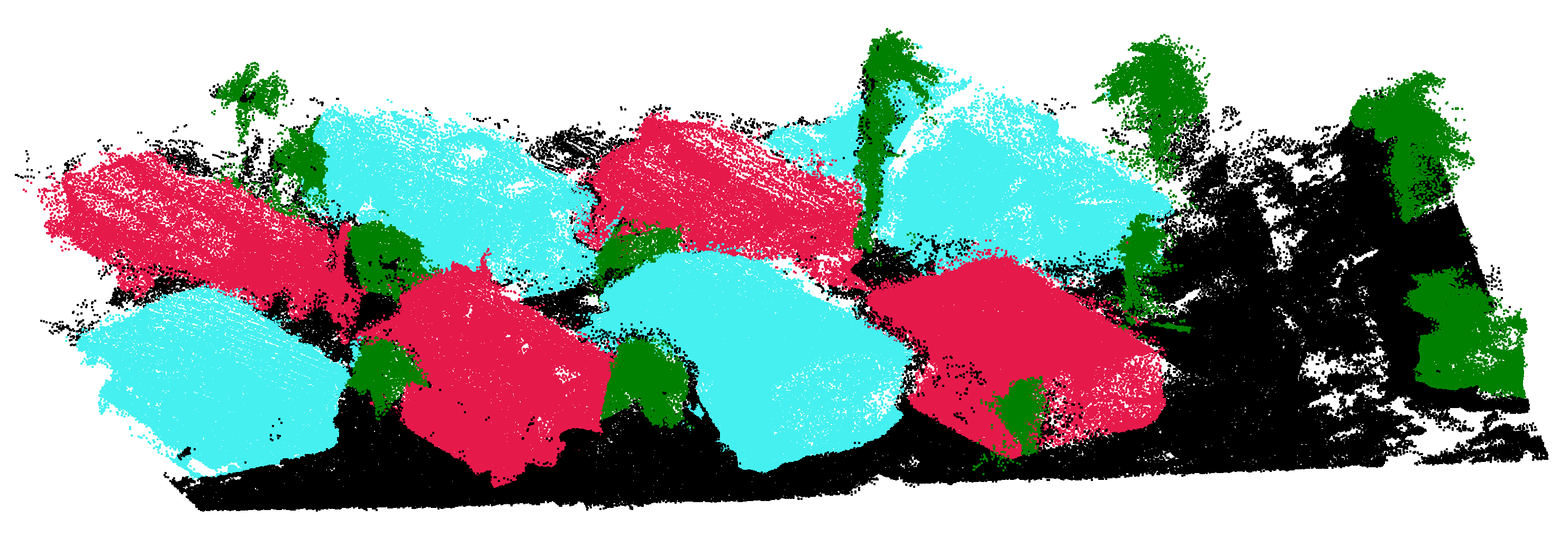} \\ \hline
\textbf{FPT}& \includegraphics[width=0.7\linewidth]{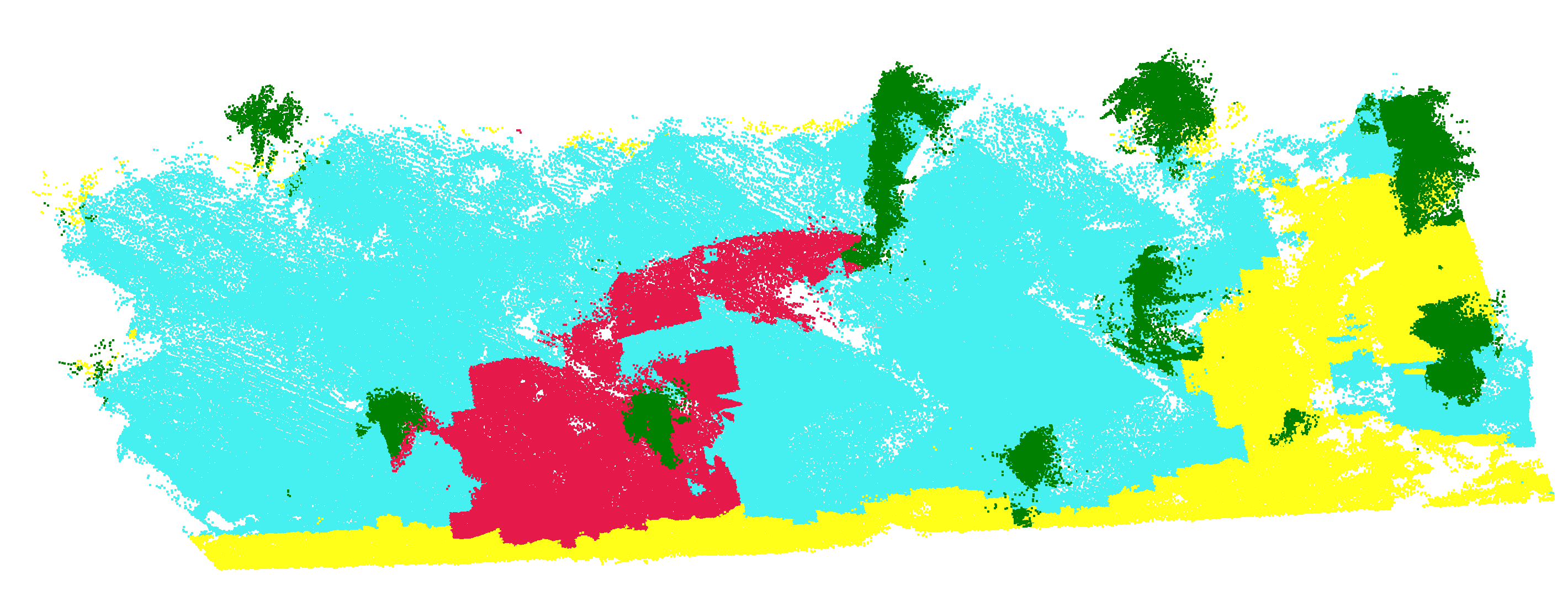} \\ \hline
\textbf{PTv3} &\includegraphics[width=0.7\linewidth]{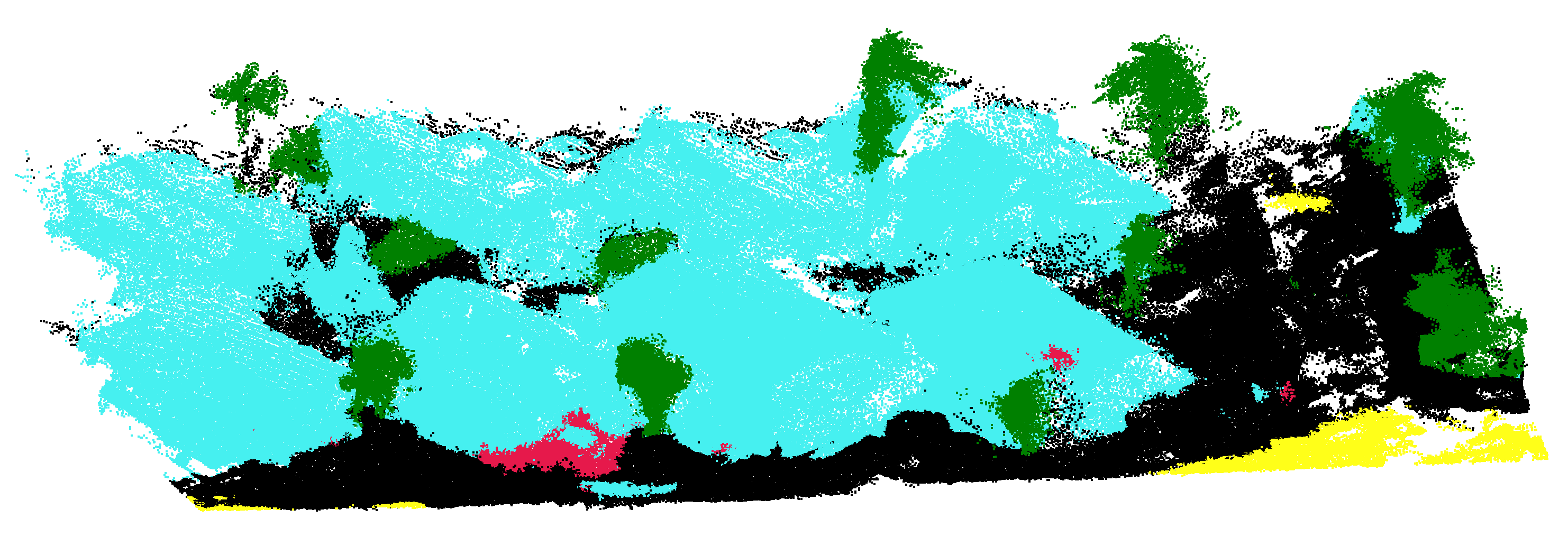}  \\ \hline
\textbf{OA-CNNs} & \includegraphics[width=0.7\linewidth]{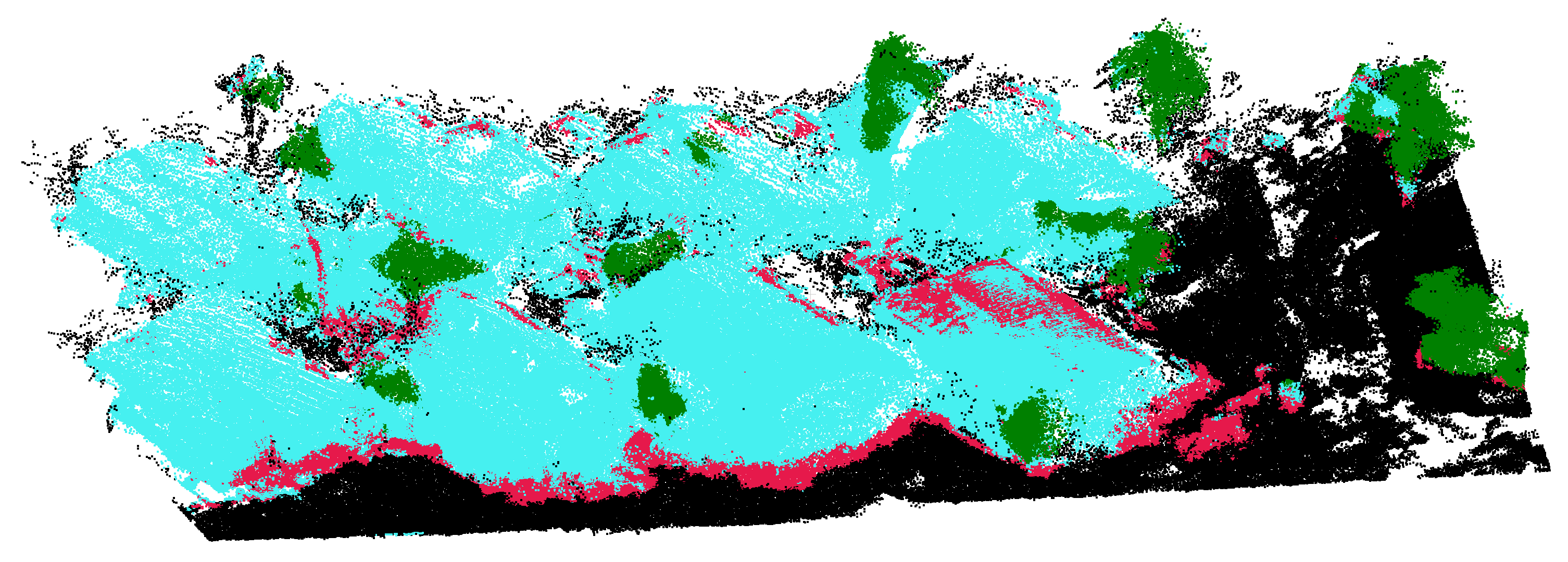} \\ \hfill
\end{tabular}
\label{tab:res3}
\end{table}
}

\section{Discussion and Conclusion}
In this study, we evaluated the performance of the SOTA 3D deep learning-based semantic segmentation methods, FPT \cite{park2022fast}, PTv3 \cite{wu2024ptv3}, and OA-CNNs \cite{Peng_2024_CVPR}, on our reconstructed 3D dataset specifically designed for post-disaster scenarios. Despite their superior performances on common indoor and outdoor 3D semantic segmentation benchmarks, our experiments revealed their limitations when applied to post-disaster environments. All models exhibited difficulty in accurately distinguishing between damaged and undamaged buildings, likely due to the subtle texture variations in post-disaster settings. These issues contributed to low overall performance metrics, with the best mean IoU (mIoU) remaining well below expectations. These findings emphasize the urgent need for developing 3D semantic segmentation models that are robust to the unique complexities of post-disaster environments. In particular, future work should focus on enhancing sensitivity to structural damage features and designing architectures that can better generalize to outdoor disaster scenarios. Furthermore, the release of high-quality, annotated 3D datasets tailored for post-disaster assessment, like the one we proposed, will be critical to fostering progress in this domain.


\small
\bibliographystyle{IEEEtranN}
\bibliography{references}

\end{document}